# CIGAN: A Python Package for Handling Class Imbalance using Generative Adversarial Networks

## Yuxiao Huang [a*] and Yan Ma [b]

[a] Data Science, Columbian College of Arts & Sciences, George Washington University, Washington, D.C., U.S.A.
[b] Department of Biostatistics and Bioinformatics, Milken Institute School of Public Health, George Washington University, Washington, D.C., U.S.A.
[*] yuxiaohuang@gwu.edu.

A key challenge in Machine Learning is class imbalance, where the sample size of some classes (majority classes) are much higher than that of the other classes (minority classes). If we were to train a classifier directly on imbalanced data, it is more likely for the classifier to predict a new sample as one of the majority classes. In the extreme case, the classifier could completely ignore the minority classes. This could have serious sociological implications in healthcare, as the minority classes are usually the disease classes (e.g., death or positive clinical test result). In this paper, we introduce a software that uses Generative Adversarial Networks to oversample the minority classes so as to improve downstream classification. To the best of our knowledge, this is the first tool that allows multi-class classification (where the target can have an arbitrary number of classes). The code of the tool is publicly available in our github repository (https://github.com/yuxiaohuang/research/tree/master/gwu/working/cigan/code).

## 1. Introduction

One of the most important applications of Machine Learning is prediction. When the target (a variable whose value we are interested in) is discrete, we call a value of the target a *Class*.

Predicting the class of a target is then referred to as *Classification*. Take the Drug Consumption (DC) dataset (https://archive.ics.uci.edu/ml/datasets/Drug+consumption+%28quantified%29) for example. The target *Amphetamine* has 3 classes: "*Never Used*" (class 1), "*Used over a Decade Ago*" (class 2) and "*Used in Last Decade*" (class 3) (Actually there are another four classes: "*Used in Last Year*", "*Used in Last Month*", "*Used in Last Week*" and "*Used in Last Day*". As they all belong to class "*Used in Last Decade*" (class 2), they were merged into this class). The classification problem is then predicting which of the above three classes a new subject belongs to.

A key challenge in classification is class imbalance, where the number of samples for some classes are much higher than those for other classes. In the DC dataset, class 1 has 976 (or 52%) samples (the majority class), whereas class 2 and 3 have 230 and 679 (or 12% and 36%) samples (the minority classes). The problem of class imbalance is that, if we were to train a model directly on imbalanced data, it would be more likely for the model to predict a sample as a majority class. In the extreme case, the model could completely ignore the minority classes and, in turn, predict each sample as the majority class. In the DC dataset, this means that the model could predict every subject as class 1 (*Never Used*). However, wrongly predicting a subject who has used drugs as never used will have profound sociological implications.

A primary approach for addressing class imbalance is over sampling, which aims to generate samples for the minority classes so as to balance the data. However, traditional over sampling methods[1-8] tend to generate minority class samples that are indistinguishable from the majority class samples, resulting in overfitting (the model trained on the balanced data does not generalize well in reality)[9].

As a deep generative neural network, Generative Adversarial Networks (GANs)[10] have been largely applied in computer vision. Recently, GANs were also extended to address class imbalance, by first learning the underlying distribution of the minority classes and then using the distribution to augment these classes[9]. Unlike traditional oversampling methods, it is much less likely for GANs to generate minority class samples that are indistinguishable from the majority class samples. It has been shown that, using GANs to augment the minority classes can significantly improve the downstream classification[9].

Despite GANs' superior performance in handling class imbalance, to the best of our knowledge, there is no off-the-shelf tool that can be widely applied. While an implementation of GANs was proposed to handle class imbalance [11] it only allows binary classification (where the target only has two classes), hence cannot be used for multi-class classification (where the target has more than two classes, which is the case in the DC dataset).

The goal of this paper is to introduce a software, named Class Imbalance GAN (CIGAN), that uses GANs to oversample the minority classes so as to improve downstream classification. It is worth noting that, the software proposed in this paper permits any arbitrary number of classes (i.e., both binary classification and multi-class classification). As a result, compared to previous work[11], this software can be applied to a wider range of problems.

The rest of the paper is organized as follows. In Section 2, we introduce the idea behind GANs. In Section 3, we demonstrate how to use CIGAN in a step-by-step fashion using the DC dataset as an example. In Section 4, we conclude the paper and discuss future work.

## 2. Method

GANs comprise two parts, a generator and a discriminator. The generator aims to *generate* data that resembles the real data, whereas the discriminator aims to *discriminate* between the real data and the generated data. Through an adversarial game between the two parts, the discriminator becomes more and more accurate in terms of separating the real data and generated data (so as to debunk the generator), whereas the generator becomes more and more accurate in terms of generating realistic data (so as to deceive the discriminator). This adversarial game eventually reaches a *Nash Equilibrium*, where the generator generates perfectly realistic data so that the discriminator can no longer distinguish the generated data from the real data.

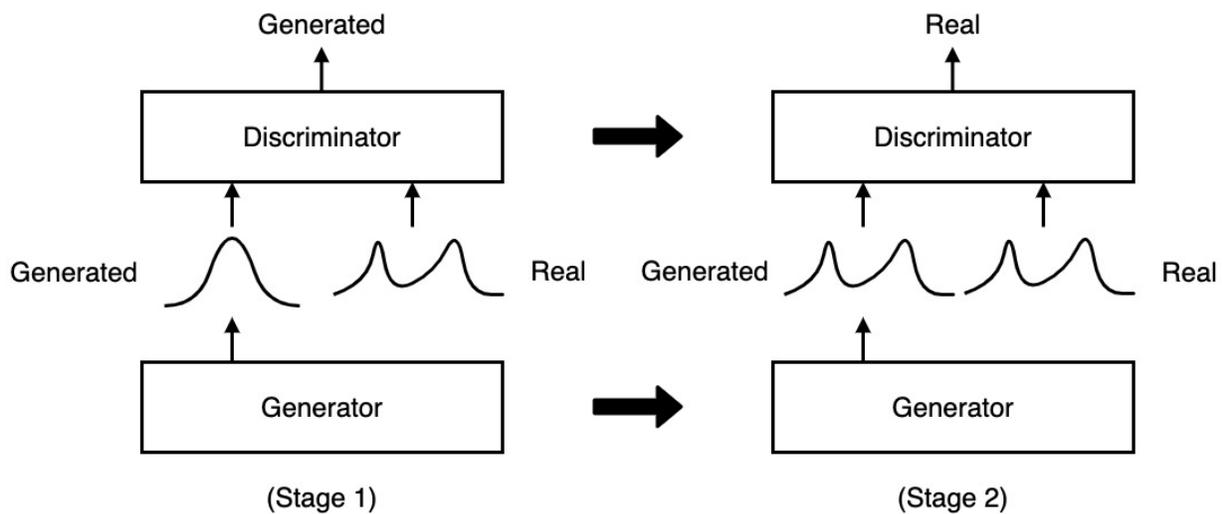

Figure 1. An oversimplified example of the adversarial game between generator and discriminator in GANs.

An oversimplified example of the adversarial game in GANs is also shown in Figure 1. For illustrative purposes, we assume GANs only take two stages to reach the Nash Equilibrium. Specifically, in stage 1 (left panel in Figure 1) the generator first generates data and then passes the data to the discriminator. The discriminator compares the data it received with the real data and claims that the received data are generated, as its distribution (unimodal, as shown in the left panel) is different from the distribution of the real data (bimodal). This insight (real data being bimodal) is passed to the generator, in stage 2 (right panel in Figure 1) the generator uses this information and generates data that is identical to the real data (as shown in the right panel) and passes the generated data to the discriminator. As the two kinds of data are indistinguishable, the discriminator claims that the received data are real.

With the example of the adversarial game in hand, we now explain how to use GANs to address the class imbalance problem. Specifically, for each minority class we want to augment, GANs first learn its latent distribution and then use the distribution to generate samples for the minority class (so that it can have the same number of samples as the majority class). Take the DC dataset for example. Suppose we would like to augment both of the minority classes, class 2 and 3. As class 1 (the majority class) has 976 samples, we will have 976 samples for both class 2 and 3 after data augmentation.

Figure 2 shows the pipeline of classification with CIGAN addressing the class imbalance problem. The pipeline includes four steps. In step 1, we preprocess the imbalanced data and split the preprocessed data into imbalanced training, validation, and test data. In step 2, we feed the training data to CIGAN, which augments the minority classes. In step 3, we first feed the augmented training data and the validation data to a classifier to train the classifier and fine-tune its hyperparameters. Next, we select the classifier that corresponds to the best hyperparameter setting (i.e., which leads to the best classification performance on the validation data). Last, we feed the test data to the best classifier to generate the test score (classification performance on the test data). See the code for the pipeline on DC dataset in our github repository (https://github.com/yuxiaohuang/research/tree/master/gwu/working/cigan/code).

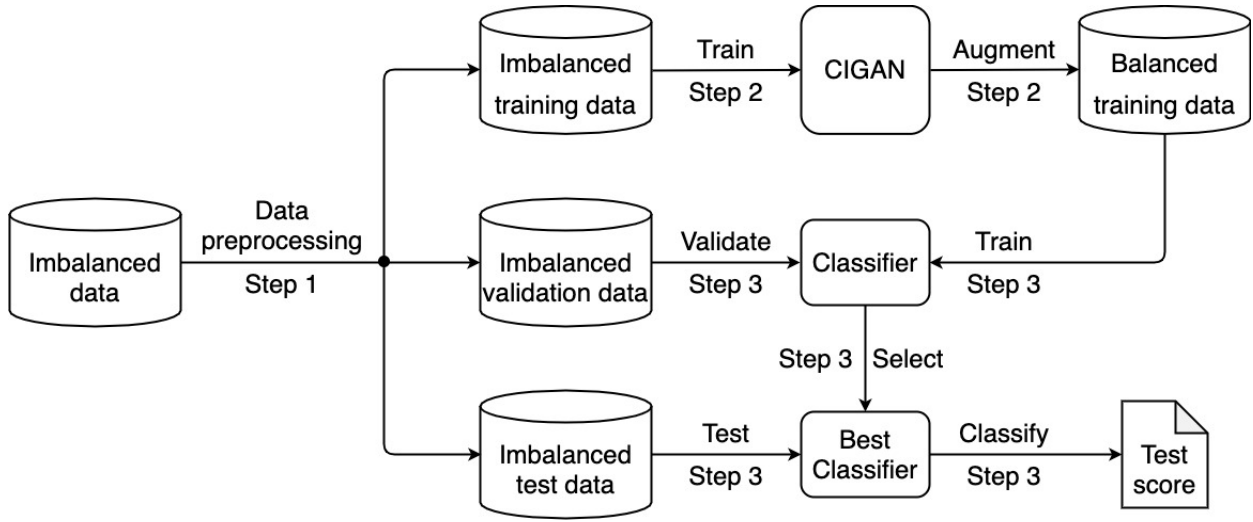

Figure 2. The pipeline of classification with CIGAN.

## 3. Software

In this section, we propose the Class Imbalance GAN (CIGAN) class, which takes as input imbalanced training data, uses GANs to augment the minority classes, and outputs balanced training data. Specifically, in section 3.1 we show the simplest way to use CIGAN for data augmentation, by using the default parameter settings of the class. This allows the reader to quickly produce balanced data and, in turn, obtain a baseline for the downstream classification. In section 3.2, we provide a detailed discussion of the parameters of CIGAN as well as some good practices for fine tuning these parameters, which could significantly boost the downstream classification. Lastly, in section 3.3, we experimentally demonstrate the performance of CIGAN on the DC dataset.

3.1. Using the default parameters settings

It is fairly straightforward to apply CIGAN for data augmentation when using its default settings. As shown in Figure 3, this only involves two lines of code. Concretely, in line 2 we declare an object (i.e., an instance) of the CIGAN class. In line 5 we use this object and its *fit_resample* function, which 1) takes as input imbalanced training data (where *X_train* includes the features and *y_train* the target), 2) uses GANs to augment all the minority classes, and 3) outputs balanced training data.

```
In [38]:  1  # The CIGAN
          2  cigan = CIGAN()
          3
          4  # Augment the minority classes in the training data
          5  X_gan_train, y_gan_train = cigan.fit_resample(X_train, y_train)
```

Figure 3. Applying the CIGAN class for data augmentation by using its default parameter settings.

3.2. Fine-tuning the parameters

While using the default parameter settings is straightforward, it may not lead to the best results (i.e., balanced data that boost the downstream classification the most). Similar to other machine learning models, we will have to fine-tune the parameters of the CIGAN class so as to get the best results. Before discussing the good practices for doing so, let us first explain the following 16 parameters in the CIGAN class:

1. *X_train*. The features in the training data.
2. *y_train*. The target in the training data.
3. *minor_classes*='all'. This parameter specifies the list of minority classes need to be oversampled. The default 'all' means we want to oversample all the minority classes. If, instead, we only want to oversample some minority classes, we can pass the list of these classes to the parameter.
4. *coding_size*='auto'. This parameter specifies the dimension (an integer) of the latent gaussian noise (which is the input of the generator in CIGAN). The default 'auto' means we will set this number as half of the number of features.
5. *batch_size*=32. This parameter specifies the batch size for minibatch gradient descent. The default is 32 (samples).

6. *max_iter*=10. This parameter specifies the maximum iteration (a.k.a., epoch) for minibatch gradient descent. The default is 10.

7. *generator_hidden_layer_sizes*=[100, 200, 300, 400, 500]. This parameter specifies the hidden layer sizes of the generator in CIGAN. The default [100, 200, 300, 400, 500] (a list of five integers) means there are five hidden layers, where the number of perceptron from layer 1 to layer 5 ranges from 100 to 500.

8. *discriminator_hidden_layer_sizes*=[500, 400, 300, 200, 100]. This parameter specifies the hidden layer sizes of the discriminator in CIGAN. The default [500, 400, 300, 200, 100] (a list of five integers) means there are five hidden layers, where the number of perceptron from layer 1 to layer 5 ranges from 500 to 100.

9. *generator_hidden_layer_activation*='selu'. This parameter specifies the activation function of the hidden layers in the generator. The default is 'selu'.

10. *discriminator_hidden_layer_activation*='selu'. This parameter specifies the activation function of the hidden layers in the discriminator. The default is 'selu'.

11. *generator_optimizer*=keras.optimizers.Adam. This parameter specifies the optimizer for the generator. The default is keras.optimizers.Adam.

12. *discriminator_optimizer*=keras.optimizers.Adam. This parameter specifies the optimizer for the discriminator. The default is keras.optimizers.Adam.

13. *generator_learning_rate*=10 ** -4. This parameter specifies the learning rate for the generator. The default is 10 ** -4.

14. *discriminator_learning_rate*=10 ** -4. This parameter specifies the learning rate for the discriminator. The default is 10 ** -4.

15. *random_seed*=42. This parameter specifies the random seed. The default is 42.

16. *n_jobs*=1. This parameter specifies the number of CPU cores used for parallelization. The default 1 means we only use one CPU (i.e., no parallelization).

With the definition of the above parameters in hand, we can now discuss the good practices for fine-tuning the key parameters. Concretely, we will begin with the parameter that could lead to the most significant change to downstream classification, and we will end with the parameter that could lead to the least significant change.

1. *generator_learning_rate* and *discriminator_learning_rate* (order 13 and 14 in the list above). If the downstream validation score is not satisfactory, we may either decrease or increase the two learning rates (e.g., by a factor of 10). It is worth noting that the two learning rates do not need to be the same.

2. *generator_hidden_layer_sizes* and *discriminator_hidden_layer_sizes* (order 7 and 8 in the list above). If the downstream validation score is not satisfactory, we may either decrease or increase the layer size (e.g., by a factor of 10). Concretely, if the data size is large, we tend to increase the layer size. Otherwise, we tend to decrease the size. It is worth noting that, as in order 7 and 8 in the list above, the *generator_hidden_layer_sizes* should be in ascending order (i.e., becoming wider and wider) whereas *discriminator_hidden_layer_sizes* should be in descending order (becoming narrower and narrower).

3. *max_iter* (order 6 in the list above). If the downstream validation score is not satisfactory, we may increase the maximum iteration (e.g., by a factor of 2).

3.3. Empirical results

In this section we use the DC dataset as an example to experimentally demonstrate that using CIGAN for data augmentation could significantly improve downstream classification. We applied the same pipeline shown in Figure 2 in our experiment. See the data, code and full results in paper github repository.

For the classifier in the pipeline, we applied Gradient Boosting Machines (GBMs), which is one of the most popular machine learning classifiers[12]. Specifically, we used the latest package of GBMs in sklearn, named HistGradientBoostingClassifier (HGBC). We measured the performance of HGBC using three of the most widely used metrics for classification performance, including precision (a.k.a., positive predictive value), recall (a.k.a., sensitivity) and F1-score (the harmonic mean of the two). All of these metrics range in value from 0 to 1 (where 0 / 1 means the model predicts every class incorrectly / correctly), and the higher the value, the better the performance of the classifier.

As shown in Table 1, CIGAN always outperform the baseline in all the metrics, across all three classes in the DC dataset (i.e., "*Never Used*", "*Used over a Decade Ago*" and "*Used in Last Decade*"). Moreover, for class *Used over a Decade Ago* (class 2), the recall of CIGAN (0.246) is more than twice as large as the recall of the baseline (0.116). This is a significant improvement considering the profound socio-economic implications of not being able to identify people who have used drugs.

Table 1. Precision, recall, F1-score of HGBC on the test data, where HGBC was trained either on the original imbalanced training data (Baseline), or on the balanced training data augmented by CIGAN. The scores in bold are the higher ones in the same column.

| Method | Precision (class 1) | Precision (class 2) | Precision (class 3) | Recall (class 1) | Recall (class 2) | Recall (class 3) | F1 (class 1) | F1 (class 2) | F1 (class 3) |
|---|---|---|---|---|---|---|---|---|---|
| Baseline | 0.647 | 0.308 | 0.636 | **0.730** | 0.116 | 0.652 | 0.686 | 0.168 | 0.644 |
| GANs | **0.686** | **0.447** | **0.639** | **0.730** | **0.246** | **0.677** | **0.707** | **0.318** | **0.657** |

## 4. Discussions

In this paper we propose CIGAN to address the class balance and, in turn, improve downstream classification. This is done by using CIGAN to augment the minority classes so that each class has the same number of samples. However, making all the classes having the same amount of samples may not be the best way to boost classification, as it may lead to overfitting. Alternatively, augmenting the minority classes so that their number of samples approaches (but not necessarily equals) the number for majority classes could better improve classification. We are currently investigating to what degree we should augment the minority classes so as to boost classification the most. We will incorporate this optimal degree into CIGAN in our future work.

## Acknowledgements

This research was supported by the National Institute on Minority Health and Health Disparities of the National Institutes of Health under Award Number R01MD013901 and the George Washington University Facilitating Fund (UFF) FY21.